\documentclass[11pt,letterpaper]{article}
\usepackage{acl2015}
\usepackage{times}
\usepackage{latexsym}
\usepackage{helvet}
\usepackage{courier}
\usepackage{amssymb}
\usepackage{amsmath,amsfonts,amssymb}
\usepackage{bbding}
\usepackage{url}
\usepackage{mathrsfs}
\usepackage{graphicx}
\usepackage{algorithmic}
\usepackage{subfigure}
\usepackage{multirow}
\usepackage{comment}
\usepackage[ruled,vlined]{algorithm2e}
\usepackage{color}
\usepackage{balance}
\usepackage[table,usenames,dvipsnames]{xcolor}

\title{Abstractive Multi-Document Summarization via Phrase Selection and Merging\thanks{~The work described in this paper is substantially supported by grants from the Research and Development Grant of Huawei Technologies Co. Ltd (YB2013090068/TH138232) and the Research Grant Council of the Hong Kong Special Administrative
Region, China (Project Codes: 413510 and 14203414). \newline The work was done when Weiwei Guo was in Columbia University}}

\author{Lidong Bing$^{\S}$ \ \ Piji Li$^{\natural}$ \ \ Yi Liao$^{\natural}$ \ \ Wai Lam$^{\natural}$\\ \textbf{Weiwei Guo}$^{\dag}$  \ \ \textbf{Rebecca J. Passonneau}$^{\ddag }$ \\
$^{\S}$Machine Learning Department, Carnegie Mellon University, Pittsburgh, PA USA\\
$^{\natural}$Department of Systems Engineering and Engineering Management, \\The Chinese University of Hong Kong\\
$^{\dag}$Yahoo Labs, Sunnyvale, CA, USA\\
$^{\ddag}$Center for Computational Learning Systems, Columbia University, New York, NY, USA\\
{\tt $^{\S}$lbing@cs.cmu.edu, $^{\natural}$\{pjli, yliao, wlam\}@se.cuhk.edu.hk}\\
{\tt $^{\dag}$wguo@yahoo-inc.com, $^{\ddag}$becky@ccls.columbia.edu}\\}

\date{}

\begin{document}
\maketitle

\begin{abstract}
We propose an abstraction-based multi-document summarization framework that can construct new sentences by exploring more fine-grained syntactic units than sentences, namely, noun/verb phrases.
Different from existing abstraction-based approaches, our method
first constructs a pool of concepts and facts represented by phrases from the input documents.
Then new sentences are generated by selecting and merging informative phrases to maximize the salience of phrases and meanwhile satisfy the sentence construction constraints.
We employ integer linear optimization for conducting phrase selection and merging simultaneously in order to achieve the global optimal solution for a summary.
Experimental results on the benchmark data set TAC 2011 show that our framework outperforms the state-of-the-art models under automated pyramid evaluation metric, and achieves reasonably well results on manual linguistic quality evaluation.
\end{abstract}

\section{\label{sec:Introduction}Introduction}

Existing multi-document summarization (MDS) methods fall in three categories: extraction-based, compression-based and abstraction-based.
Most summarization systems adopt the \textbf{extraction-based} approach which selects some original sentences
from the source documents to create a short summary~\cite{Erkan:2004:LGL:1622487.1622501,Wan:2007:MBT:1625275.1625743}.
However, the restriction that the whole sentence should be selected potentially yields some overlapping information in the summary. To this end, some researchers apply compression on the selected sentences by deleting words or phrases \cite{Knight00statistics-basedsummarization,Lin:2003:ISP:1118935.1118936,Zajic06sentencecompression,Harabagiu:2010:UTT:1777432.1777436,ra-mds:piji:2015}, which is the {\bf compression-based} method.
Yet, these compressive summarization models cannot merge facts from different source sentences, because all the words in a summary sentence are solely from one source sentence.

\begin{figure*}[!t]
\centering
\includegraphics[width=5.6in,height=2.6in]{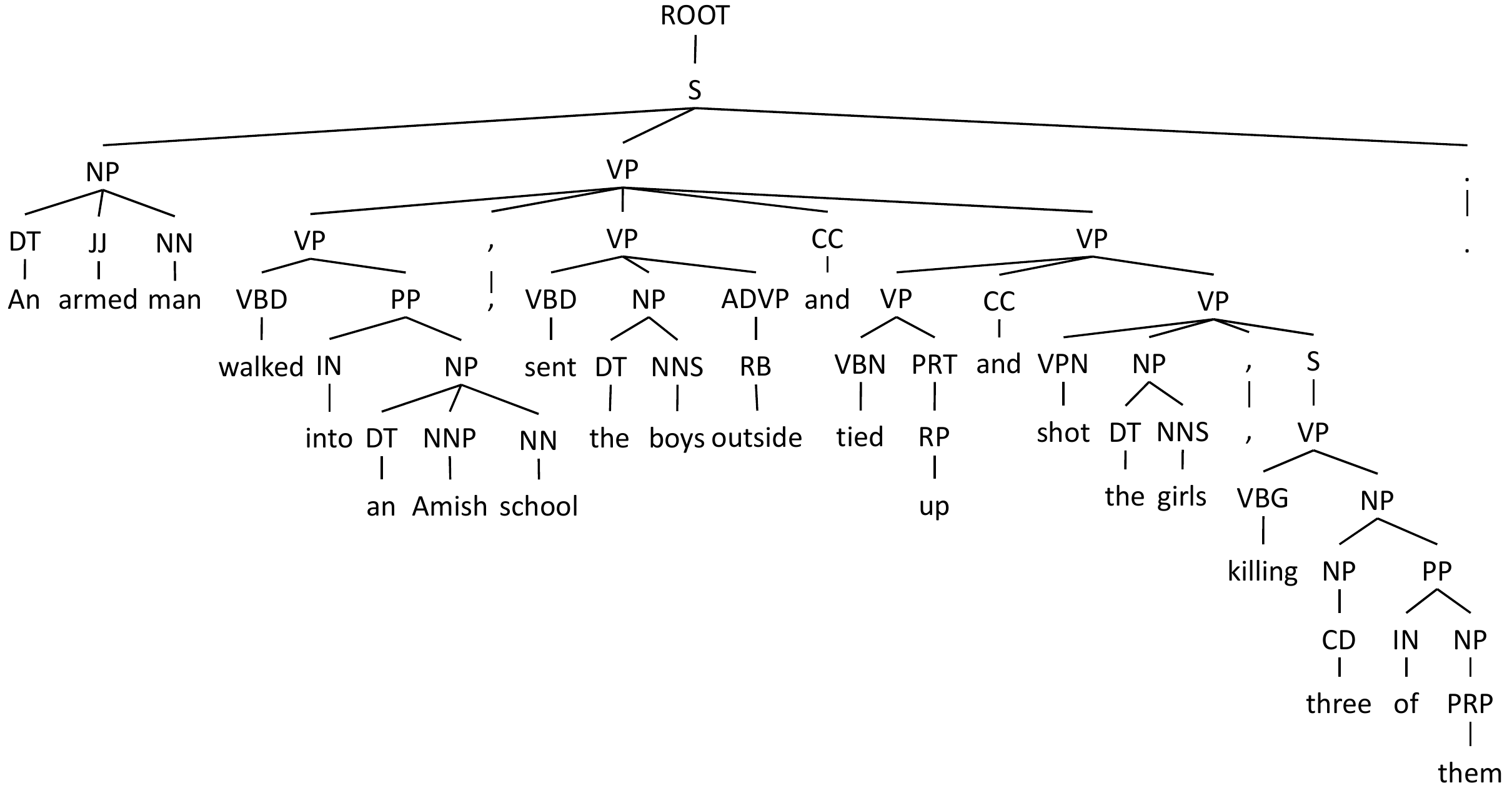}
\caption{\label{fig:syntactic_tree}The constituency tree of a sentence from a news document.}
\end{figure*}

In fact, previous investigations show that human-written summaries are more abstractive, which can be regarded as a result of sentence aggregation and fusion \cite{cheung-penn-2013a,Jing:2000:CPB:974305.974329}.
Some works, albeit less popular, have studied \textbf{abstraction-based} approach that can
construct a sentence whose fragments come from different source sentences. One important work developed by
Barzilay and McKeown~\shortcite{Barzilay:2005:SFM:1108994.1108996} employed
sentence fusion, followed by \cite{Filippova:2008:SFV:1613715.1613741,Filippova:2010:MCF:1873781.1873818}.
These works first conduct clustering on sentences to compute the salience of topical themes.
Then, sentence fusion is applied within each cluster of related sentences to generate
a new sentence containing common information units of the sentences.  The abstractive-based approaches gather information across sentence boundary, and hence have the potential to cover more content in a more concise manner.


%

In this paper, we propose an abstractive MDS framework that can construct new sentences by exploring
more fine-grained syntactic units than sentences,
namely, noun/verb phrases (NPs/VPs). This idea is based on two observations.
First, the major constituent phrases
loosely correspond to the concepts and facts. After reading a set of documents
describing the same topic or event, a person digests these documents as key concepts and facts in his/her mind, such as ``{\it an armed man}'' and ``{\it walked into an Amish school}'' from Figure \ref{fig:syntactic_tree}.
Second, a summary writer re-organizes the key concepts and facts 
to form new sentences for the summary. 
Accordingly, our proposed framework has two major
components corresponding to the above observations. The first component
creates a pool of concepts and facts represented by NPs and VPs from the input documents.
A salience score is computed for each phrase by exploiting redundancy of the document content in a global manner.
The second component constructs new sentences by selecting and merging phrases based on their salience scores, and ensures the validity of new sentences using a integer linear optimization model.

The contribution of this paper is two folds. (1) We extract NPs/VPs from constituency trees to represent key concepts/facts, and merge them to construct new sentences, which allows more summary content units (SCUs) \cite{conf/naacl/NenkovaP04} to be included in a sentence by breaking the original sentence boundaries.
(2) The designed optimization framework for addressing the problem is unique and effective. Our optimization algorithm {\bf simultaneously} selects and merges a set of phrases that maximize the number of covered SCUs in a summary. Meanwhile, since the basic unit is phrases, we design compatibility relations among NPs and VPs, as well as other optimization constraints, to ensure that the generated sentences contain correct facts.
Compared with the sentence fusion approaches that compute salience scores of sentence clusters, our proposed framework explores a more fine-grained textual unit (i.e., phrases), and maximizes the salience of selected phrases in a global manner.


\section{Description of Our Framework}
We first introduce how to extract NPs and VPs from constituency trees, and subsequently calculate salience scores for them. Then we formulate the sentence generation task as an optimization problem, and design constraints. In the end, we perform several post-processing steps to improve the order and the readability of the generated sentences.
%


\subsection{\mbox{Phrase Salience Calculation}}
The first component decomposes the sentences in documents into a set of noun phrases (NPs) derived from the subject parts of a constituency tree and a set of verb-object phrases (VPs), representing potential key concepts and key facts, respectively.  These phrases will serve as the basic elements for sentence generation.

We employ Stanford parser \cite{Klein:2003:AUP:1075096.1075150}
to obtain a constituency tree for each input sentence. After that,
we extract NPs and VPs from the tree as follows: (1) The NPs and VPs that are the direct
children of the sentence node (represented by the \textbf{S} node) are extracted.
(2) VPs (NPs) in a path on which all the nodes are VPs (NPs) are also recursively extracted
and regarded as having the same parent node \textbf{S}.
Recursive operation in the second step will only be carried out in two levels
since the phrases in the lower levels may not be able to convey a complete fact.
Take the tree in Figure~\ref{fig:syntactic_tree} as an example, the corresponding sentence is decomposed into
phrases ``{\it An armed man}'', ``{\it walked into an Amish school, sent the boys outside and tied up and shot the girls, killing three of them}'', ``{\it walked into an Amish school}'', ``{\it sent the boys outside}'', and ``{\it tied up and shot the girls, killing three of them}''. \footnote{
We only consider the recursive operation for a VP with more than one
parallel sub-VPs, such as the highest VP in Figure~\ref{fig:syntactic_tree}.
The sub-VPs following modal, link or auxiliary verbs are not extracted
as individual VPs. In addition, we also extract the clauses functioning as subjects of
sentences as NPs, such as ``that clause''. Note that we also mention such clauses
as ``noun phrase'' although their syntactic labels could be ``SBAR'' or ``S''.}
Because of the recursive operation, the extracted phrases may have overlaps.
Later, we will show how to avoid such overlapping in phrase selection.


A salience score is calculated for each phrase to indicate its importance.
Different types of salience can be incorporated in our framework,
such as position-based method \cite{Yih:2007:MSM:1625275.1625563},
statistical feature based method \cite{Woodsend:2012:MAS:2390948.2390978},
concept-based method \cite{PKUTM-2011}, etc.
One key characteristic of our approach is that the considered basic units are phrases
instead of sentences.
Such finer granularity leaves more room for better global salience score by potentially covering more distinct facts.
In our implementation, we adopt a concept-based weight
incorporating the position information. The concept
set is designated to be the union set of unigrams, bigrams, and named entities in
the documents. We remove stopwords
and perform lemmatization before extracting unigrams and bigrams.
The position-based term frequency
is used in the concept weighting scheme.
When counting the frequency, each occurrence of a concept in an input
document is weighted with the paragraph position.
The weight larger than 1 is given to the concept occurrences in the first
few paragraphs. Specifically, the weight of the first paragraph is
$B$ and the weight decreases as the position of the paragraph increases from the beginning of the document.
The weighting function is:
\begin{equation} \label{}
 H(p) =
\left\{
\begin{aligned}
    \begin{array}{cc}
               \rho^p*B & \mbox{if }p < -(\log{B}/\log{\rho}) \\
               1  & \mbox{otherwise}
             \end{array}
\end{aligned}
\right.,
\end{equation}
where $p$ is the position of the paragraph starting from 0, from beginning of the document, and $\rho$ is a positive constant and smaller than 1.
Then, the salience of
a phrase is calculated as the summed weights of its concepts.









\subsection{New Sentence Construction Model}
The construction of new sentences is formulated as an optimization problem
which is able to simultaneously generate a group of sentences.
Each new sentence is composed of one NP and at least one VP, where the NP and VPs may come from different source sentences.
In the process of new sentence generation, the compatibility relation between NP and VP and a variety of summarization requirements are jointly considered.

\subsubsection{Compatibility Relation}
Compatibility relation is designed to indicate whether an NP and a VP can be used to form a new sentence.
For example, the NP ``{\it Police}" from another sentence
should not be the subject of the VP ``{\it sent the boys outside}" extracted from Figure \ref{fig:syntactic_tree}. We use some heuristics to find compatibility, and then expand the compatibility relation to more phrases by extracting coreference.

To find coreference NPs (different mentions for the same entity), we first conduct coreference resolution
for each document with Stanford coreference resolution package~\cite{Lee:2013:DCR:2576217.2576221}.
We adopt those resolution rules that are able to achieve high quality
and address our need for summarization. In particular,
Sieve 1, 2, 3, 4, 5, 9, and 10 in the package are used.
A set of clusters are obtained and each cluster contains
the mentions that refer to the same entity in a document.
The clusters from different documents in the same topic are merged
by matching the named entities.
After merging, the mentions that are not NPs extracted in the
phrase extraction step are removed in each cluster. Two NPs
in the same cluster are determined as alternative of each other.

To find alternative VPs, Jaccard Index is employed as
the similarity measure. Specifically, each VP is represented as
a set of its concepts and the index value is calculated for
each pair of VPs. If the value is larger than a threshold, the two VPs
are determined as alternative of each other.


We then define an indicator matrix $\Gamma_{|\textbf{N}||\textbf{V}|}$,
in which $\Gamma[i,j]=1$ if an NP $N_i$ and a VP $V_j$ come from the
same node \textbf{S} in the constituency tree, otherwise, $\Gamma[i,j]=0$.
Let $\tilde{\textbf{N}}_i$ and $\tilde{\textbf{V}}_i$ represent
the alternative phrases of $N_i$ and $V_i$ as described above.
The compatibility matrix $\tilde{\Gamma}_{|\textbf{N}||\textbf{V}|}$ is
defined as follows: 
\begin{equation} \label{e:compatibility}
\tilde{\Gamma}[p,q] = \begin{cases}
  1 & \mbox{if } N_p\in\tilde{\textbf{N}}_i \wedge \Gamma[i,q]=1 \\
   1 &  \mbox{if } V_q\in\tilde{\textbf{V}}_j \wedge \Gamma[p,j]=1 \\
 1 & \mbox{if } \Gamma[p,q]=1 \\
0  & \mbox{otherwise} \end{cases}
\end{equation}
where $\tilde{\Gamma}[p,q]=1$
means $N_p$ and $V_q$ are compatible/permitted for constructing a new sentence.
$\tilde{\Gamma}$ is the final compatibility matrix that we use in the optimization.
The first case of Equation~\ref{e:compatibility} implies that
if $N_p$ and $N_i$ are coreferent, $N_p$ can replace
$N_i$ and serve as the subject of $N_i$'s VP (i.e., $V_q$).
The second case implies that if $V_q$ is very similar to $V_j$,
$V_q$ can be concatenated to $V_j$'s NP (i.e., $N_p$).

\subsubsection{Phrase-based Content Optimization}
The overall objective function of our optimization formulation to select NPs and VPs is defined as: 
\begin{equation}
\label{e:objective}
\begin{split}
\max\{ & \sum_i{\alpha_i S^N_i} - \sum_{i<j}{\alpha_{ij}(S^N_i+S^N_j)R^N_{ij}}  \\
 & +\sum_i{\beta_i S^V_i} - \sum_{i<j}{\beta_{ij}(S^V_i+S^V_j)R^V_{ij}}\},
\end{split}
\end{equation}
where $\alpha_i$ and $\beta_i$ are selection indicators for the NP $N_i$ and the VP $V_i$, respectively.
$S^N_i$ and $S^V_i$ are the salience scores of  $N_i$ and $V_i$.
$\alpha_{ij}$ and $\beta_{ij}$ are co-occurrence indicators of pairs
($N_i$, $N_j$) and ($V_i$, $V_j$).
$R^N_{ij}$ and $R^V_{ij}$ are the similarity of pairs
($N_i$, $N_j$) and ($V_i$, $V_j$). If $N_i$ and $N_j$
are coreferent, $R^N_{ij}=1$. Otherwise, the similarity is calculated with
the above Jaccard Index based method.
The notations are summarized in Table~\ref{t:notation}.

Specifically, we maximize the salience score of the selected NPs and VPs
as indicated by the first and the third terms in Equation~\ref{e:objective},
and penalize the selection of similar NP pairs and similar VP pairs as
indicated by the second and the fourth terms.
Meanwhile, the phrase selection is governed by
a set of constraints so that the selected phrases can generate valid sentences.
The constraints will be explained in details in Section \ref{sec:constraints}.

One characteristic of our objective function is that
NPs and VPs are treated differently, i.e., there are different selection/penalty terms for NP and VP.
Such design enables us to avoid the false penalty between an NP and a VP.
For example, the algorithm produces two sentences: the first sentence is ``{\it the gunman shot ...}" with an NP
``{\it the gunman}'', and the other sentence has a VP ``{\it confirmed the gunman died}''. Obviously, we should not penalize the redundancy between them, because mentioning the gunman is necessary in both sentences.

\begin{table}[!t]
\centering
\begin{small}
\begin{tabular}{@{}p{1.4cm}@{~}|@{~}p{6.2cm}@{}}
  \hline
Notation & Description \\
  \hline
$N_i$, $V_i$   &  Noun phrase $i$ and verb phrase $i$   \\
$\alpha_i$, $\beta_i$  &  Selection indicators of $N_i$ and $V_i$ \\
$\alpha_{ij}$, $\beta_{ij}$ &  Co-occurrence indicators of pairs ($N_i$, $N_j$) and ($V_i$, $V_j$) \\
$S^N_i$, $S^V_i$  &  Salience scores of $N_i$ and $V_i$ \\
$R^N_{ij}$, $R^V_{ij}$  &  Similarity of pair ($N_i$, $N_j$) and pair ($V_i$, $V_j$) \\
  \hline
  \hline
$\Gamma_{|\textbf{N}||\textbf{V}|}$  &  $\Gamma[i,j]=1$ if $N_i$ and $V_j$ are from the same sentence \\
$\tilde{\textbf{N}}_i$, $\tilde{\textbf{V}}_i$  &  The alternative phrases of $N_i$ and $V_i$ \\
$\tilde{\Gamma}_{|\textbf{N}||\textbf{V}|}$  &  $\tilde{\Gamma}[i,j]=1$ means $N_i$ and $V_j$ are
compatible for being used to construct a new sentence \\
$\tilde{\gamma}_{ij}$  &  Sentence generation indicator for $N_i$ and $V_j$ if $\tilde{\Gamma}[i,j]=1$ \\
  \hline

  \hline
\end{tabular}
\caption{\label{t:notation}Notations.}
\end{small}
\end{table}

\subsubsection{Sentence Generation Constraints}
\label{sec:constraints}

To summarize the related sentences in the documents, human writers usually merge the
important facts in different VPs about the same entity into a single sentence,
and omit the trivial facts.
Also, the same entity is likely to be described by coreferent NPs.
Therefore, in our approach, only one NP is selected and employed as the subject
of the newly generated sentence, which is then concatenated with the merged facts (i.e., VPs).
If the compatibility entry $\tilde{\Gamma}[i,j]$
for $N_i$ and $V_j$ is 1, we define a sentence generation indicator
$\tilde{\gamma}_{ij}$ to indicate whether
both $N_i$ and $V_j$ are selected to construct a new sentence in the summary.

We design the following groups of constraints to realize our aim of
phrase selection and new sentence construction.  The objective function and constraints are linear, therefore the problem can be solved by existing Integer Linear Programming (ILP)
solvers such as simplex algorithm~\cite{Dantzig:1997:LPI:248375}.

\noindent\textbf{\texttt{NP validity}}.
To maintain the consistency between the selection indicator $\alpha$ and the compatibility entry $\tilde{\Gamma}$ for NP $N_i$, we introduce two constraints as follows:  
  \begin{equation}\label{e:np_validity}
\forall i, j, \alpha_i\geq \tilde{\gamma}_{ij}; ~~ \forall i, \sum_{j}{\tilde{\gamma}_{ij}}\geq \alpha_i.
\end{equation}
These two constraints work together to ensure the valid assignment of $\alpha$ according to the compatibility entry $\tilde{\Gamma}$.

\noindent\textbf{\texttt{VP legality}}.
Similarly, the following requirement guarantees the consistency between the selection indicator $\beta$ and the compatibility entry $\tilde{\Gamma}$ for selected VP $V_i$: 
\begin{equation}\label{e:vp_validity}
\forall j, \sum_{i}{\tilde{\gamma}_{ij}} = \beta_j.
\end{equation}

\noindent The above two constraints
jointly ensure that the selected NPs and VPs are able to form new summary sentences according to the values of sentence generation indicators.

\noindent\textbf{\texttt{Not i-within-i}}.
Two phrases in the same path of a constituency tree cannot be chosen at the same time:  
  \begin{equation}\label{}
    \begin{array}{c}
\mbox{if } \exists{V_k\rightsquigarrow V_j}, \mbox{then } \beta_k+\beta_j \leq 1, \\
\mbox{if } \exists{N_k\rightsquigarrow N_j}, \mbox{then } \alpha_k+\alpha_j \leq 1.
\end{array}
\end{equation}
For example, ``{\it walked into an Amish school, sent the boys outside and tied up and shot the girls, killing three of them}''
and ``{\it walked into an Amish school}'' cannot be both incorporated in the summary, because of the obvious redundancy.

\noindent\textbf{\texttt{Phrase co-occurrence}}.
These constraints control the co-occurrence relation of NPs or VPs. 
For NPs, we introduce three constraints:
\begin{eqnarray}
\label{e:co_alpha_1} \alpha_{ij}-\alpha_{i}\leq 0,  \\
\label{e:co_alpha_2} \alpha_{ij}-\alpha_{j}\leq 0,  \\
\label{e:co_alpha_3} \alpha_{i}+\alpha_{j}-\alpha_{ij}\leq 1.
\end{eqnarray}
Constraints ~\ref{e:co_alpha_1} to ~\ref{e:co_alpha_3} ensure a valid solution of NP selection.
The first two constraints state that if
the units $N_i$ and $N_j$ co-occur in the summary (i.e., $\alpha_{ij}=1$), 
then we have to include them individually (i.e., $\alpha_{i}=1$ and $\alpha_{j}=1$).
The third constraint is the inverse of the first two.
Similarly, the constraints for VPs are as follows:
\begin{eqnarray}
\label{e:co_beta_1} \beta_{ij}-\beta_{i}\leq 0,  \\
\label{e:co_beta_2} \beta_{ij}-\beta_{j}\leq 0,  \\
\label{e:co_beta_3} \beta_{i}+\beta_{j}-\beta_{ij}\leq 1.
\end{eqnarray}



\noindent\textbf{\texttt{Sentence number}}.
In abstractive summarization, we do not prefer to generate
  many short sentences. This is
  controlled by:
\begin{equation}\label{}
\sum_{i}{\alpha_{i}}\leq K,
\end{equation}
where $K$ is the maximum number of sentences.


\noindent\textbf{\texttt{Short sentence avoidance}}.
We do not select the VPs from very short sentences because
    a short sentence normally cannot convey a complete key fact \cite{Woodsend:2012:MAS:2390948.2390978}.
  \begin{equation}\label{}
    \mbox{if } l(\textbf{S}) < M, V_i \in \textbf{S}, \mbox{then }{\beta_i=0},
\end{equation}
where $M$ is the threshold of the sentence length.

\noindent\textbf{\texttt{Pronoun avoidance}}.
We exclude the NPs that are pronouns from being
  selected as the subject of the new sentences. As
  previously observed \cite{Woodsend:2012:MAS:2390948.2390978},
 pronouns are normally not used by human summary writers.
  It is because the summary is short and the narration relation
  of sentences is relatively simple so that pronouns are not needed.
  Moreover, in automatic summary, pronouns will cause ambiguity in the
  summary, especially when the sentence order is automatically
  determined. Therefore, we model the constraint as:

 \begin{equation}\label{}
   \mbox{if } N_i~is~pronoun, \mbox{then } \alpha_i = 0.
\end{equation}

%

\noindent\textbf{\texttt{Length constraint}}. The overall length of the selected
  NPs and VPs is no larger than a limit $L$:
  \begin{equation}\label{}
    \sum_{i}\{l(N_i)*\alpha_i\} + \sum_{j}\{l(V_j)*\beta_j\}\leq L,
\end{equation}
where $l()$ is the word-based length of a phrase.



\subsection{Postprocessing}
Recall that we require that one NP and at least one VP compose a sentence.
Thus, we form a raw sentence with a selected NP as the subject followed by
the corresponding selected VPs that are indicated by sentence generation indicator
$\tilde{\gamma}_{ij}$ having the value 1.
The VPs in a summary sentence are ordered according to
their natural order if they come from the same document.
Otherwise, they are ordered according to the timestamps of the corresponding documents.
After that, if the total length is smaller than $L$,
we add conjunctions such as ``and'' and ``then'' to concatenate the VPs for improving the readability
of the newly generated sentences.
The pseudo-timestamp of a sentence is defined as the earliest timestamp of its VPs
and the sentences are ordered based on their pseudo-timestamps.

\subsection{Relation to Existing MDS Approaches}
Many existing extraction-based and
compression-based MDS approaches could be regarded as special cases under our framework:
(1) To simulate extraction-based summarization,
we just need to constrain that the highest NP and the highest VP from the same sentence are
selected simultaneously. In addition, no NPs and VPs in lower levels can be selected.
Thus, the output only contains the original sentences of the source documents.
(2) To simulate compression-based summarization, we can adapt
our framework to conduct sentence selection and sentence compression in a joint manner.
Specifically, we only need to restrict that the NP and VPs of a summary sentence
must come from the same original sentence.

\section{Experiments}

\subsection{Experimental Setup}

The data set of traditional summarization task in Text Analysis Conference (TAC) 2011 is used to evaluate
the performance of our approach. This data set is the latest one and it contains
44 topics. Each topic falls into one of 5 predefined
event categories and contains 10 related news documents.
There are four writers to write model summaries for each topic.

The data set of traditional summarization task in TAC 2010 is employed as the development/tuning
data set. This data set contains 46 topics from
the same predefined categories. Each topic also has 10 documents and 4 model summaries.

Based on the tuning set, the key parameters of our model are set as follows.
The constants $B$ and $\rho$ in the weighting function are set to 6 and 0.5 repectively.
The similarity threshold in obtaining the alternative VPs is 0.75.
We did not observe significant difference between cosine similarity and Jaccard Index.

We mainly evaluate the system by pyramid evaluation.
To gain a comprehensive understanding, we also evaluate by ROUGE evaluation and manual linguistic quality evaluation.



\subsection{Results with Pyramid Evaluation}
The pyramid evaluation metric \cite{conf/naacl/NenkovaP04} involves semantic matching
of summary content units (SCUs) so as to recognize alternate realizations of the same meaning.
Different weights are assigned to SCUs based on their frequency in model summaries.
A weighted inventory of SCUs named a pyramid
is created, which constitutes a resource
for investigating alternate realizations of the same meaning.
Such property makes pyramid method more suitable to evaluate summaries.
Another widely used evaluation metric is ROUGE~\cite{Lin:2003:AES:1073445.1073465}
and it evaluates summaries from word overlapping perspective. Because
of the strict string matching, it ignores the semantic content units and performs better when
larger sets of model summaries are available. In contrast
to ROUGE, pyramid scoring is robust with as
few as four model summaries \cite{conf/naacl/NenkovaP04}.
Therefore, in recent summarization evaluation workshops such as TAC, the pyramid
is used as the major metric.

\begin{table}[!t]
\centering
\begin{tabular}{c|c|c|c}
  \hline
 & Auto-pyr & Auto-pyr & Rank in \\
System & (Th: .6)   & (Th: .65) & TAC 2011 \\
\hline
Our  & 0.905 & 0.793 & NA \\
22  & 0.878 &  0.775 & 1 \\
43 & 0.875 &  0.756  & 2 \\
17  & 0.860 & 0.741  & 3 \\
  \hline
\end{tabular}
\caption{\label{t:pyramid_compare} Comparison with the top 3 systems in TAC 2011.}
\end{table}

Since manual pyramid evaluation is time-consuming, and the exact evaluation scores are not reproducible
especially when the assessors for our results are different from those of TAC,
we employ the automated version of pyramid proposed in \cite{Passonneau:13}.
The automated pyramid scoring procedure relies on distributional semantics to assign
SCUs to a target summary.  Specifically, all n-grams within sentence bounds are extracted,
and converted into 100 dimension latent topical vectors via a weighted matrix
factorization model \cite{Guo:2012:MSL:2390524.2390644}. Similarly, the contributors and
the label of an SCU are transformed into 100 dimensional vector representations.
An SCU is assigned to a summary if there exists an n-gram such
that the similarity score between the SCU low dimensional vector and the
n-gram low dimensional vector exceeds a threshold.
Passonneau et al.~\shortcite{Passonneau:13}
showed that the distributional similarity based method produces automated
scores that correlate well with manual pyramid scores, yielding more accurate
pyramid scores than string matching based automated methods \cite{harnly_automation_2005}.
In this paper, we adopt the same setting as in
\cite{Passonneau:13}: a 100 dimension matrix factorization model is learned
on a domain independent corpus, which is drawn from sense definitions of WordNet and
Wiktionary\footnote{http://en.wiktionary.org/}, and Brown corpus.
We experiment with 2 threshold values, i.e., 0.6 and 0.65, similar to those used in \cite{Passonneau:13}.

The top three systems in TAC 2011 evaluated with manual pyramid score
were System 22~\cite{PKUTM-2011}, 43, and 17~\cite{Ng_swing:exploiting}. Table~\ref{t:pyramid_compare} shows
the comparison with them under the automated pyramid evaluation.
Our method achieves the best results in both thresholds, which means that our method is able to find more semantic content units (SCUs) than the state-of-the-art system in TAC 2011.
In addition, paired t-test (with  $p<0.01$) comparing our model with the
best system in TAC 2011, i.e., System 22, shows that the performance of our model is significantly better.
It is worth noting that the three systems used additional external linguistic resources: System 22 used a Wikipedia corpus for providing domain knowledge,
System 17 and 43 defined some category-specific features. Without any domain adaption, our framework can still achieve encouraging performance.


We calculate Pearson's correlation
to measure how well the automatic pyramid approximates the manual pyramid scores
for 50 system submissions in TAC 2011.
The values are 0.91 and 0.93 for thresholds 0.6 and 0.65 respectively.
It demonstrates that the automated pyramid is reliable to
differentiate the performance of different methods.


\begin{table}[!t]
\centering
\begin{tabular}{@{}c@{~}|@{~}c@{~}|@{~}c@{~}|@{~}c@{~}|@{~}c@{~}|@{~}c@{~}|@{~}c@{}}
  \hline
 &  \multicolumn{3}{c}{\textbf{ROUGE-2}} &  \multicolumn{3}{c}{\textbf{ROUGE-SU4}} \\
\hline
 System & P & R & F1 & P & R & F1\\
 \hline
Our & 0.117   &  0.117   &  0.117   &  0.148   &  0.147   &  0.148 \\
22 &  0.112   &  0.114   &  0.113   &  0.147   &  0.150   &  0.148 \\
43  & 0.132   &  0.135   &  0.134   &  0.162   &  0.166   &  0.164 \\
17 &  0.128   &  0.131   &  0.129   &  0.157   &  0.160   &  0.159  \\
  \hline
\end{tabular}
\caption{\label{t:rouge_compare} Performance under ROUGE metric.}
\end{table}

\subsection{Results with ROUGE Evaluation}

As mentioned above, we favor the pyramid evaluation over the ROUGE score because
it can measure the summary quality beyond simply string matching. Here, we also provide ROUGE
score for our reference.
ROUGE-1.5.5 package\footnote{http://www.berouge.com/Pages/default.aspx}
is employed with the same parameters as in TAC.
The results are summarized in Table~\ref{t:rouge_compare}.
Our performance is slightly better than System 22, and it is not as good as System 43 and 17.
The reason is that System 43 and 17 used category-specific features and
trained the feature weights with the category information in TAC 2010 data.
These features help them select better
category-specific content for the summary. However, the usability of such features depends on the
availability of predefined categories in the summarization task, as well as the availability
of training data with the same predefined categories for estimating feature weights.
Therefore, the adaptability of these methods is limited to some extent. In contrast,
our framework does not define any category-specific feature and only
uses TAC 2010 data to tune the parameters for general summarization purpose.

\subsection{Linguistic Quality Evaluation}
The linguistic quality of summaries is evaluated using the five
linguistic quality questions on grammaticality (Q1), non-redundancy (Q2), referential
clarity (Q3), focus (Q4), and coherence (Q5) in Document Understanding Conferences (DUC).
A Likert scale with five levels is employed with 5 being very good with 1 being very poor.
A summary was blindly evaluated by three assessors on each question.
System 22 performed better than System 43 and 17 in TAC 2011 on the evaluation of readability, which is an aggregation
of the above questions. Considering the intensive labor force of manual
assessment, we only conduct comparison with System 22.

The results are given in Table~\ref{t:linguisti_quality}.
On average, the two systems perform very closely.
System 22 is an extraction-based method that picks the original sentences, hence
it achieves higher score in Q1 grammaticality, while our approach has
some new sentences with grammar mistakes, which is a common
problem for abstractive methods and deserves more future research effort.
For Q4 focus, our score is higher than System 22, which reveals that
our summary sentences are relatively more cohesive. The score of Q3 referential
clarity shows that the
referential relation is basically clear in our summaries, even when
new sentences are automatically generated.
In general, ignoring the grammaticality scores, our system still performs better than System 22.
Specifically, the average scores of our system and System 22 on the last
four questions are 3.37 and 3.33 respectively.

\begin{table}[!t]
\centering
\begin{tabular}{c|@{~}c@{~}|@{~}c@{~}|@{~}c@{~}|@{~}c@{~}|@{~}c@{~}|@{~}c}
\hline
System  & Q1 & Q2 & Q3 & Q4 & Q5 & AVG\\
 \hline
Our  &  3.67  & 	3.50  & 	3.90  & 	3.23  & 	2.83  & 	3.43 \\
22 & 4.13  & 	3.50  & 	3.97  & 	2.97  & 	2.87  & 	3.49 \\
  \hline
\end{tabular}
\caption{\label{t:linguisti_quality}Evaluation of linguistic quality.}
\end{table}

%

\section{Qualitative Results}

\subsection{Analysis of Summary Sentence Type}
There are three types of sentences in the summaries generated by our
framework, namely, new sentences, compressed sentences, and original sentences.
A new sentence is constructed by merging the phrases from different original sentences.
A compressed sentence is generated by deleting phrases from an original sentence.
An original sentence in the summary is directly extracted from the input documents.

The percentage of different types of sentences in our summaries is calculated.
About 33\% of the summary sentences are newly constructed. This demonstrates that
our framework has good capability of merging phrases from the original sentences so as to
convey more information in compacted summaries. In addition, about 44\% of the summary sentences
are generated by compression.
It shows a unique characteristic of our framework: sentence construction and sentence compression are conducted in a unified model.

\subsection{Case Study}
Table~\ref{t:amish_summary} shows the summary of the first topic, i.e., ``{\it Amish Shooting}'', by our framework.
The summary sentence ID and the sentence type are given in
the form of ``\texttt{[summary sentence ID: sentence type]}''.
Each selected phrase and the original sentence ID
where the phrase originated are given in the form of ``\{\texttt{selected phrase (original sentence ID)}\}''.
There are three compressed sentences
with IDs 1, 2, and 4, one new sentence with ID 3, and two original sentences with IDs 5 and 6.

The new sentence is constructed from the following original sentences
in which the extracted NPs and VPs are indicated with colored parentheses:

\vspace{0.2cm}
\noindent{
\begin{tabular}{@{}p{7.7cm}@{}}
  \hline
   \textcolor[rgb]{0.50,0.00,0.50}{\texttt{(84)}}: On Monday morning, \textcolor[rgb]{0,1,0}{\texttt{(NP}} Charles Carl Roberts IV\textcolor[rgb]{0,1,0}{\texttt{)}} \textcolor[rgb]{0,1,0}{\texttt{(VP}} \textcolor[rgb]{0,0,1}{\texttt{(VP}} entered the West Nickel Mines Amish School in Lancaster County\textcolor[rgb]{0,0,1}{\texttt{)}} and \textcolor[rgb]{0,0,1}{\texttt{(VP}} shot 10 girls\textcolor[rgb]{0,0,1}{\texttt{)}}, \textcolor[rgb]{0,0,1}{\texttt{(VP}} killing five\textcolor[rgb]{0,0,1}{\texttt{)}}\textcolor[rgb]{0,1,0}{\texttt{)}}. \\
   \textcolor[rgb]{0.50,0.00,0.50}{\texttt{(85)}}: \textcolor[rgb]{0,1,0}{\texttt{(NP}} Roberts\textcolor[rgb]{0,1,0}{\texttt{)}} \textcolor[rgb]{0,1,0}{\texttt{(VP}} killed himself as police stormed the building\textcolor[rgb]{0,1,0}{\texttt{)}}. \\
   \textcolor[rgb]{0.50,0.00,0.50}{\texttt{(150)}}: \textcolor[rgb]{0,1,0}{\texttt{(NP}} Roberts\textcolor[rgb]{0,1,0}{\texttt{)}} \textcolor[rgb]{0,1,0}{\texttt{(VP}} left what they described as rambling notes for his family\textcolor[rgb]{0,1,0}{\texttt{)}}.\\
  \hline
 \vspace{1mm}
\end{tabular}
}
The NPs of these sentences are coreferent so that some of their VPs are
merged and concatenated with one NP, i.e., ``{\it Charles Carl Roberts IV}''.

The summary sentences with IDs 1, 2, and 4 are compressions from
the following original sentences respectively:

\vspace{0.2cm}
\noindent{
\begin{tabular}{@{}p{7.7cm}@{}}
  \hline
   \textcolor[rgb]{0.50,0.00,0.50}{\texttt{(25)}}: \textcolor[rgb]{0,1,0}{\texttt{(NP}} An armed man\textcolor[rgb]{0,1,0}{\texttt{)}} \textcolor[rgb]{0,1,0}{\texttt{(VP}}\textcolor[rgb]{0,0,1}{\texttt{(VP}} walked into an Amish school\textcolor[rgb]{0,0,1}{\texttt{)}}, \textcolor[rgb]{0,0,1}{\texttt{(VP}} sent the boys outside\textcolor[rgb]{0,0,1}{\texttt{)}} and \textcolor[rgb]{0,0,1}{\texttt{(VP}} tied up and shot the girls, killing three of them\textcolor[rgb]{0,0,1}{\texttt{)}}\textcolor[rgb]{0,1,0}{\texttt{)}}, \textcolor[rgb]{0,1,0}{\texttt{(NP}} authorities\textcolor[rgb]{0,1,0}{\texttt{)}} \textcolor[rgb]{0,1,0}{\texttt{(VP}} said\textcolor[rgb]{0,1,0}{\texttt{)}}. \\
   \textcolor[rgb]{0.50,0.00,0.50}{\texttt{(64)}}: \textcolor[rgb]{0,1,0}{\texttt{(NP}}\textcolor[rgb]{0,0,1}{\texttt{(NP}} A man\textcolor[rgb]{0,0,1}{\texttt{)}}who laid siege to a one-room Amish schoolhouse\textcolor[rgb]{0,1,0}{\texttt{)}},\textcolor[rgb]{0,1,0}{\texttt{(VP}} killing five girls\textcolor[rgb]{0,1,0}{\texttt{)}},\textcolor[rgb]{0,1,0}{\texttt{(VP}}\textcolor[rgb]{0,0,1}{\texttt{(VP}} told his wife shortly before opening fire that he had molested two young girls who were his relatives decades ago\textcolor[rgb]{0,0,1}{\texttt{)}}and\textcolor[rgb]{0,0,1}{\texttt{(VP}} was tormented by ``dreams of molesting again''\textcolor[rgb]{0,0,1}{\texttt{)}}\textcolor[rgb]{0,1,0}{\texttt{)}},\textcolor[rgb]{0,1,0}{\texttt{(NP}} authorities\textcolor[rgb]{0,1,0}{\texttt{)}}\textcolor[rgb]{0,1,0}{\texttt{(VP}} said Tue\textcolor[rgb]{0,1,0}{\texttt{)}}. \\
   \textcolor[rgb]{0.50,0.00,0.50}{\texttt{(145)}}: According to media reports, \textcolor[rgb]{0,1,0}{\texttt{(NP}} the gunman\textcolor[rgb]{0,1,0}{\texttt{)}} \textcolor[rgb]{0,1,0}{\texttt{(VP}}\textcolor[rgb]{0,0,1}{\texttt{(VP}} was not Amish\textcolor[rgb]{0,0,1}{\texttt{)}} and \textcolor[rgb]{0,0,1}{\texttt{(VP}} had not attended the school\textcolor[rgb]{0,0,1}{\texttt{)}}\textcolor[rgb]{0,1,0}{\texttt{)}}.\\
  \hline
   \vspace{1mm}
\end{tabular}
}
Some uncritical information is excluded from the summary sentences, such as ``{\it sent the boys outside}'',
``{\it authorities said}'', etc.
In addition, the VP ``{\it killing five girls}'' of the original sentence with ID 64 is also excluded since it has significant redundancy
with the summary sentence with ID 1.

\begin{table}[!t]
\centering
\begin{tabular}{@{}p{7.7cm}@{}}
  \hline
 \textcolor[rgb]{0.00,0.00,1.00}{\texttt{[\textbf{1}:C]}} \{An armed man \textcolor[rgb]{0.50,0.00,0.50}{\texttt{(25)}}\} \{walked into an Amish school \textcolor[rgb]{0.50,0.00,0.50}{\texttt{(25)}}\} \{tied up and shot the girls, killing three of them. \textcolor[rgb]{0.50,0.00,0.50}{\texttt{(25)}}\}
 \textcolor[rgb]{0.00,0.00,1.00}{\texttt{[\textbf{2}:C]}} \{A man who laid siege to a one-room Amish schoolhouse \textcolor[rgb]{0.50,0.00,0.50}{\texttt{(64)}}\} \{told his wife shortly before opening fire that he had molested two young girls who were his relatives decades ago \textcolor[rgb]{0.50,0.00,0.50}{\texttt{(64)}}\} \{was tormented by dreams of molesting again.	 \textcolor[rgb]{0.50,0.00,0.50}{\texttt{(64)}}\}
 \textcolor[rgb]{0.00,0.00,1.00}{\texttt{[\textbf{3}:N]}} \{Charles Carl Roberts IV \textcolor[rgb]{0.50,0.00,0.50}{\texttt{(84)}}\} \{killed himself as police stormed the building， \textcolor[rgb]{0.50,0.00,0.50}{\texttt{(85)}}\}
\{left what they described as rambling notes for his family. \textcolor[rgb]{0.50,0.00,0.50}{\texttt{(150)}}\}
 \textcolor[rgb]{0.00,0.00,1.00}{\texttt{[\textbf{4}:C]}} \{The gunman \textcolor[rgb]{0.50,0.00,0.50}{\texttt{(145)}}\} \{was not Amish	 \textcolor[rgb]{0.50,0.00,0.50}{\texttt{(145)}}\} \{had not attended the school. \textcolor[rgb]{0.50,0.00,0.50}{\texttt{(145)}}\}
 \textcolor[rgb]{0.00,0.00,1.00}{\texttt{[\textbf{5}:O]}} \{The shootings \textcolor[rgb]{0.50,0.00,0.50}{\texttt{(148)}}\} \{occurred about 10:45 a.m.\textcolor[rgb]{0.50,0.00,0.50}{\texttt{(148)}}\}
 \textcolor[rgb]{0.00,0.00,1.00}{\texttt{[\textbf{6}:O]}} \{Police \textcolor[rgb]{0.50,0.00,0.50}{\texttt{(149)}}\} \{could offer no explanation for the killings. \textcolor[rgb]{0.50,0.00,0.50}{\texttt{(149)}}\} \\
  \hline
\end{tabular}
\caption{\label{t:amish_summary} The summary of ``{\it Amish Shooting}'' topic.}
\end{table}

\section{\label{sec:Related}Related Work}
Existing multi-document \mbox{summarization (MDS)} works can be classified into three categories: extraction-based approaches, compression-based approaches, and abstraction-based approaches. 
%

Extraction-based approaches are the most studied of the three.
Early studies mainly followed a greedy strategy in sentence selection
\cite{DBLP:conf/icassp/CelikyilmazH11,Goldstein:2000:MSS:1117575.1117580,Wan:2007:MBT:1625275.1625743}. Each sentence
in the documents is firstly assigned a salience score.
Then, sentence selection is performed by greedily selecting the sentence
with the largest salience score among the remaining
ones. The redundancy is controlled during the selection by penalizing the remaining ones
according to their similarity with the selected sentences.
An obvious drawback of such greedy strategy is that
it is easily trapped in local optima. Later, unified models are
proposed to conduct sentence selection and redundancy control
simultaneously \cite{McDonald:2007:SGI:1763653.1763720,Filatova:2004:FMI:1220355.1220412,Yih:2007:MSM:1625275.1625563,Gillick_theicsi,Lin:2010:MSV:1857999.1858133,conf/uai/LinB12,Sipos:2012:LLS:2380816.2380846}.
However, extraction-based approaches are unable to
evaluate the salience and control the redundancy
on the granularity finer than sentences. Thus, the selected sentences
may still contain unimportant or redundant phrases.

Compression-based approaches have been investigated
to alleviate the above limitation. As a natural extension of the
extractive method, the early works adopted
a two-step approach \cite{Lin:2003:ISP:1118935.1118936,Zajic06sentencecompression,Gillick:2009:SGM:1611638.1611640}.
The first step selects the sentences, and the second step removes the unimportant or redundant units from the sentences.
Recently, integrated models have been proposed that jointly conduct sentence extraction
and compression \cite{Martins:2009:SJM:1611638.1611639,Woodsend:2010:AGS:1858681.1858739,almeida-martins:2013:ACL2013,Berg-Kirkpatrick:2011:JLE:2002472.2002534,ra-mds:piji:2015}. Note that our model also jointly conducts phrase selection and phrase merging (new sentence generation).
Nonetheless, compressive methods are unable to merge the related facts from different sentences.


On the other hand, abstraction-based approaches can generate new sentences based on the facts from different source sentences.
In addition to the previously mentioned sentence fusion work,
new directions have been explored.
Researchers developed an information extraction based approach that extracts
{\em information items} \cite{Genest:2011:FAS:2107679.2107687} or {\em abstraction schemes} \cite{Genest:2012:FAA:2390665.2390745}
as components for generating sentences.
Summary revision was also investigated to improve the quality of automatic summary
by rewriting the noun phrases or people references in the summaries
\cite{DBLP:conf/ijcnlp/Nenkova08,Siddharthan:2011:ISD:2077692.2077699}.
Sentence generation with word graph was applied
for summarizing customer opinions and chat conversations \cite{Ganesan:2010:OGA:1873781.1873820,mehdad-carenini-ng:2014:P14-1}.


 Recently, the factors of information certainty and timeline
 in MDS task were explored \cite{ng-EtAl:2014:P14-1,Wan:2014:CEM:2600428.2609559,Yan:2011:ETS:2009916.2010016}.
 Researchers also explored some variants of the typical
 MDS setting, such as query-chain focused summarization
 that combines aspects of update summarization and query-focused summarization \cite{baumel-cohen-elhadad:2014:P14-1},
 and hierarchical summarization that scales up MDS to summarize a large set of documents \cite{christensen-EtAl:2014:P14-1}.
 A data-driven method for mining sentence structures on large news archive
 was proposed and utilized to summarize unseen news events \cite{pighin-EtAl:2014:P14-1}. Moreover, some works \cite{liu2012query,kaageback2014extractive,denil2014modelling,cao2015ranking} utilized deep learning techniques to tackle some summarization tasks.

\section{Conclusions and Future Work}
We propose an abstractive MDS framework that constructs new sentences by exploring
more fine-grained syntactic units, namely, noun phrases and verb phrases.
The designed optimization framework operates on the summary level so that more complementary semantic content units can be incorporated. The phrase selection and merging is done simultaneously to achieve global optimal. Meanwhile, the constructed sentences should satisfy the constraints related to summarization requirements such as NP/VP compatibility.
Experimental results on TAC 2011 summarization data set show that our framework outperforms
the top systems in TAC 2011 under the pyramid metric.
For future work, one aspect is to enhance the grammar quality of the generated new sentences and compressed sentences.
Another aspect is to improve time efficiency of our framework, and its major bottleneck is the time consuming ILP optimzation.

\bibliographystyle{acl}
\balance
\bibliography{reference}  

\end{document}